# IMAGE BASED CHARACTER RECOGNITION, DOCUMENTATION SYSTEM TO DECODE INSCRIPTION FROM TEMPLE

Dr. Velmathi G
*School of Electronics and Communication Engineering*
*Vellore Institute of Technology Chennai*
Chennai - 600127
velmathi.g@vit.ac.in

Shangavelan M
*School of Electronics and Communication Engineering*
*Vellore Institute of Technology Chennai*
Chennai - 600127
shangavelan.m2020@vitstudent.ac.in

Harish D
*School of Electronics and Communication Engineering*
*Vellore Institute of Technology Chennai*
Chennai - 600127
harish.d@vitstudent.ac.in

Krithikshun M S
*School of Electronics and Communication Engineering*
*Vellore Institute of Technology Chennai*
Chennai - 600127
krithikshun.ms@vitstudent.ac.in

*Abstract*—This project undertakes the training and analysis of optical character recognition (OCR) methods applied to 10th-century ancient Tamil inscriptions discovered on the walls of the Brihadeeswarar Temple. The chosen OCR methods include Tesseract, a widely-used OCR engine,using modern ICR techniques to pre-process the raw data and a box editing software to finetune our model.

The Tamil language, with a rich history spanning several centuries, has witnessed significant evolution, resulting in an expanded and diversified character set over time. This research project is dedicated to enhancing optical character recognition techniques specifically tailored for the ancient Tamil script on the 10th century. The emphasis is on refining OCR methodologies while meticulously curating a comprehensive dataset specific to the intricacies of the ancient Tamil script.

Modern Tamil has over 12 'uyir eluthukal' and 18 'mei eluthukkal' and a special character. The combination of the above two yeid 216 compound characters which totals as (12+18+216+1)247 total Tamil characters,the same is applicable to 10th century tamil too.

The analysis with Tesseract aims to evaluate their effectiveness in accurately deciphering the nuances of the ancient Tamil characters. The performance of our model for the dataset are determined by their accuracy rate where the evaluated dataset divided into training set and testing set. By addressing the unique challenges posed by the script's historical context, this study seeks to contribute valuable insights to the broader field of OCR, facilitating improved preservation and interpretation of ancient inscriptions.

*Index Terms*—OCR,ICR,Tesseract

## I. INTRODUCTION

Optical Character Recognition (OCR) technology has completely changed the way ancient texts are digitized enabling researchers to avail themselves of and analyze trove of information. However, OCR for ancient scripts presents formidable challenges, particularly when dealing with languages such as 10th-century Tamil. This article explores bottlenecks encountered in OCR for 10th-century Tamil inscriptions which include lack of suitable datasets, distinctiveness of stone inscriptions and morphological changes in Tamil script over time. Consequently, we present several strategies and methodologies for enhancing accuracy and efficiency of ancient Tamil texts' OCR systems.

One key obstacle to OCR for 10th-century Tamil inscriptions is the limited collection of appropriate data sets. Unlike contemporary languages that have huge digital resources, 10th century historical Tamil literature remains extremely scarce often taking the form of fragmentary inscriptions on rocks only. Due to this lack of sufficient data set, it becomes difficult to train and develop OCR models specifically designed for ancient Tamil scripts.

The nature of 10th-century Tamil inscriptions, primarily carved into stone surfaces, presents distinct challenges for OCR. Unlike printed texts, stone inscriptions lack standardized punctuation and spacing between words, rendering conventional OCR techniques ineffective. The absence of these visual cues complicates the segmentation of text into meaningful units, necessitating innovative approaches to text recognition and reconstruction.

The Tamil writing system has changed significantly over time with characters evolving in their appearance and structure. The script found in inscriptions, from the century differs greatly from Tamil showcasing older letter shapes and intricate connections between letters. These changes in appearance create a challenge for OCR systems that are trained on contemporary Tamil leading to accuracy and reliability in recognizing the text.

In summary using OCR for deciphering 10th century Tamil inscriptions poses difficulties to limited data availability, the unique nature of stone inscriptions and the evolution of the Tamil script. Overcoming these obstacles will require collaboration across fields such, as history, linguistics and computer science. It will also involve developing specialized OCR algorithms and enhancing existing datasets. Modern Epigraph readers face many difficulties while trying to read ancient tamil inscriptions due to their degradation. The Tamil Nadu archaeology department has estimated that there are over 25000 incriptions out of which only about 10000 have been properly recognised. Out of these 25000 inscriptions about 16000 of them are from the period of interest of our project.This period was ruled by many popular chola kings such as sri.Rajaraja Chola,Rajendra Chola I,Rajendra Chola I. UNder their rule ,their kingdom had acheived many tremendous breakthroghs in art,literature,architecture,etc. Huge bronze monuments,majestic temples are some of the examples. The content and quality of these ancient inscriptions denote that the country had attained very high levels of education and wisdom. The texts that are imprinted on the temples of today were most likely written by royal poets and etched into stone by royal artisans. By conquering the challenges listed out in [1] and the degradation of inscriptions over time OCR technology can unlock a treasure trove of knowledge preserved in Tamil inscriptions contributing to our understanding of Tamil history, language and culture.

| Modern tamil | Ancient tamil in stone inscriptions(represented using modern tamil letters) | explaination |
|---|---|---|
| கே | கெ | No difference in writing down 'எ' and 'ஏ' sounds |
| கோ | கொ | No difference in writing down 'ஒ' and 'ஓ' sounds |
| மூ,மு | மு,மு | Same letter refered for bot hof these different letters.Which needs to be identified based on context in ancient tamil |
| கா | கர | Absence of a dedicated letter to represent 'அ' sounds.Instead the letter 'ர' is used and the sound is guessed from context |
| முதலாம் இராஜராஜ சோழன் | முதலாமிராஜராஜ சோழன | Absence of a word break between each words |
| ன் | ன | Absence of use of punctuations and quotations in the scriptures |

Fig. 1. Difficulties encountered when compared to modern tamil

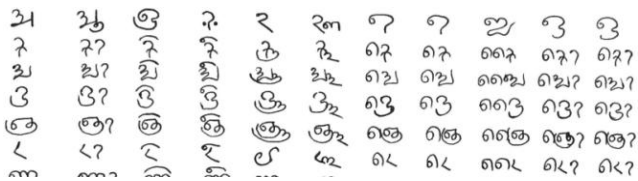

Fig. 2. Example of the scripture used in 10th century BCE

This research initiative aims to tackle these obstacles by use of OCR technology; Tesseract—an open source OCR tool commonly used for image to text conversion. Furthermore this study seeks to compile a dataset sourced from the Brihadeeswarar Temple— a repository of Tamil texts—thereby improving the authenticity and variety of training data available.

Through a comparative analysis of these two models, we aim to evaluate their effectiveness, accuracy, and applicability in digitizing ancient Tamil scriptures.

## II. MATERIALS AND METHODS

### A. Data collection

As such, to capture and subsequently digitize the inscriptions from the walls of temples a novel and novel unique approach was used by incorporating limestone powder into the process of data collection. More precisely, the incorporation of limestone powder was conducted by sprinkling a layer of limestone powder over the inscriptions on the temple wall that would significantly enhance the contrast and visibility of the letter inscriptions without actually damaging the wall in any way possible . In "Using Limestone Powder as a Novel, Non-Intrusive Data Collection Element to Enhance the Image-Based Character Recognition of Letter Inscriptions on Temple Walls" , it is highlighted that through the subtle layer, the recesses and certain contours of the character would be thereby highlighted, which would subsequently be displayed against the background.The image was captured using specialized camera equipment mounted on stable stands, with uniform lighting illuminating the entire image.Close to a 400 images where taken.Additionally images were collected from Tamil Nadu Archeological Departmart. Further with the help of various Tamil scholars in u.v.sa library we were able to gather the meaning and grammer used in ancient Tamil literature,This information was pivotal to training the dataset accurately.further To help future researchers a blog consisting of the character representation and the grammers used was compiled.

### B. Preprocessing

One of the key steps in the process of extracting meaningful information from temple inscriptions consists of preprocessing the images generated during the capturing stage. Due to the complexity of character forms, particular optimization is needed for more readable texts to secure accurate digitalization.Opencv is a huge open-source library for computer vision, machine learning, and image processing.nowadays,opencv has grown a lot that it helps us in many dynamic operations .opencv is mostly used to process videos,images,extract features from the above two,help in face recognition,or even help in pre-processing handwritten texts.etc .Image processing is the process of changing the characteristics or extracting charateristics from a image or video to help in identifying,enhancing them to be used in other APIs or for human understanding of the image. For our project we have used 2d matrix to represent images.In this format image is defined as a 2d function f(x,y),where x and y are the Cartesian coordinates and the functional value at these points given by the matrix is the intensity of grey scale value in each pixel of the image. In another words, an image is nothing but an 2d matrix with each A(i,j) to be the intensity of grey scale light at that pixel, This value id then processed as required to obtain what colour that particular pixel might be in. The image preprocessing pipeline applied comprises several key operations. The first step encompasses resizing, which adjusts the dimensions to a constant proportion, ensuring its effectiveness and easiness for subsequent computational performance .Image resizing invloves the change in scaling of the image. When the images are scaled there will be loss of details or complexity to the image thus causing the number of pixels in the image to reduce. After that, the conversion to grayscale is executed to reduce the number of dimensions by converting color changes to relevant pixel intensity contrasts.Import OpenCV and read the original image using imread() then convert to grayscale using cv2.cvtcolor() function. To implement greyscale conversion we would need to import opencv and then use cv2.cvtcolor on the image.

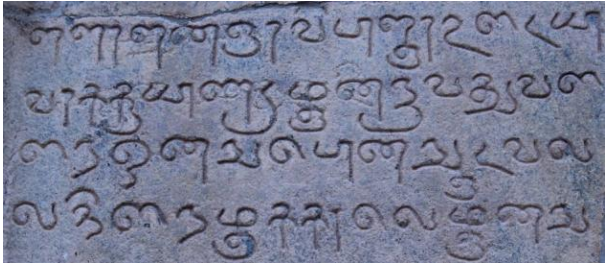

Fig. 3. Raw image

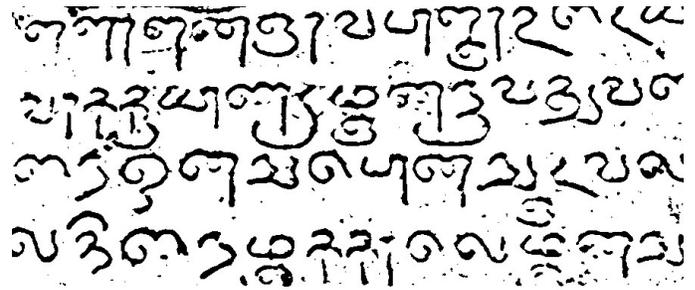

Fig. 4. OSTU Thresholding algorithm

- **Image skew and adjust size :** Bounding box position and orientation is detected and the image is rotated to follow upright position.The algorithm proposed for image skew correction begins by iterating over a range of angles, calculating a score for each angle by rotating the binary image and computing the variance between adjacent histogram values along the vertical axis. The angle yielding the highest score is selected as the optimal rotation angle. Using this angle, a rotation matrix is generated, and the original image is rotated accordingly. This process ensures automatic correction of skew in the image, making it suitable for applications such as document processing and optical character recognition (OCR).
- **OSTU Thresholding :** To deal with variations in lights and shading, an adaptive thresholding technique is hired, dynamically adjusting the brink for binarization.Image thresholding is used to assign a 2 values pixel value to each pixel in the image based on the intensity of the pixel when in greyscale format. This greyscale image and a constant called threshold constant is fed into the algorithm to obtain our binarised output.If the greyscale value of a particular pixel is greater than the threshold value then htat pixel is assigned a value of 1(background) else if the pixel vaue is lesser than the threshold value the n its assigned a value of 0(foreground).A serius problem with this method is each part of the image may need different threshold values to preserve their information properly.i.e A low threshold value when the average pixel values are high would lead to all pixels becoming the background.One of the popularly used auto thresholding algorithms are ostu thresholding.

Fig. 5. Binarised image

Most auto thresholding algorithms use the following steps,

- pre process the image into greyscale images
- Obtain the distribution of the pixel values in the image in the form of a histogram curve.
- Utilise the distribution to compute localised threshold values
- Perform usual thresholding in localised regions of the image using the calculated threshold value

The ostu algorithm is summarised in image [4]. This binary image serves as the foundation for next operations. Recognizing the ability effect of border artifacts on the evaluation, a specialized border elimination algorithm is carried out, correctly isolating the inscription content material from any surrounding disturbances. Window size and constant were found using experimentation on the images used and was manually determined. Gaussian thresholding was employed after numerous testing. Otsu's algorithm first finds intra-variances between pixels in a localised window and then tries to minimise this variance value(depends on the average of pixel intensity) by finding appropriate threshold value or pixel average.

- **MedianBlur :** The image has too much noise that have high color strength,via a median blur operation, the image is moderately blurred to lighten the effect of single pixel and small island pixel noises ensuring that handiest enormous capabilities are retained for in addition processing.The median blur uses convolution with a base kernel to get desired blurred effect. Each pixel is convolved with a ones matrix(divided by number of neighbouring pixels) essentially taking the average if the neighbouring pixels and substituting this value as the new pixel value.This tries to preserve the edges while blurring out/erasing the noise.Taking the weighted average of the neighbours introduces a effect of smudging or overlapping of pixel values in a localised group of pixels. The ones matrix/kernel mentioned above is also called the box filter.

- **Histogram equalisation :** To further enhance contrast, highlight diffused details and balance the high contrast and low brightness regions of the image, histogram equalization is added, developing a extra balanced distribution of pixel intensities. This method usually increases the global contrast of many images, especially when the usable data of the image is represented by close contrast values. Through this adjustment, the intensities can be better distributed on the histogram. This allows for areas of lower local contrast to gain a higher contrast. When the image's pixel values are represented by close contrast values, this strategy typically improves the global contrast of numerous images. The intensities of the pixels on the histogram can be more evenly spread by making this adjustment. This makes it possible for regions with less local contrast to become more contrast with each other.Histogram equalization accomplishes this by distributing the most frequently occuring intensity value over the localised window. This image is usefull in picking out the foreground from background in images where both of them have close pixel values.

  In opencv this can be implemented using cv2.equalizeHist(). Its input is the binarized image.

- **Bilateral filtering :** Aims to remove the noises in the images while keeping the edges sharp,Thus improving the quality of the image.Bilateral smoothing is an edge-preserving denoising technique.In the Gaussian smoothing technique, a weighted sum of all the pixel values in the kernel data is calculated, and the central element of the kernel is replaced with that value. But this is a function of space alone. It is not considered if the pixel lies on edge or not. This is why the Gaussian smoothing technique tends to blur out the boundaries also.Bilateral filtering or Bilateral smoothing technique overcomes this disadvantage by introducing another Gaussian filter that considers the variation of intensities to preserve the edges. Bilateral filtering can be implemented in OpenCV using the cv2.bilateralFilter() function, which takes the following parameters

  1) src: Image which is to be smoothened
  2) d: Dimension of the kernel
  3) sigmaColor: Standard deviation that controls the influence of pixels with different intensity values
  4) sigmaSpace: Standard deviation that controls the influence of distant pixels

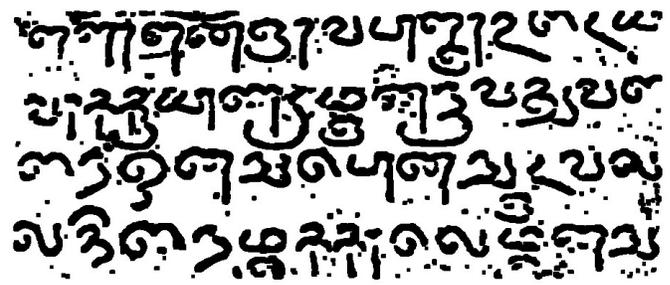

Fig. 6. Before erosion

- **Erosion :** Resulting image have very bold characters with line width.This is decreased using erosion techniques to make it better represent the base character. cv2.erode() method is used to perform erosion on the image. The basic idea of erosion is similar to soil erosion,It ships away the boundaries of the foreground and adds it into the background essentially eliminating any less concentrated pixel groups. It is normally performed on binary images. It requires 2 inputs one is our binarised image, second parameter called the kernel which decides what kind of operation is about to be performed. A pixel in the original image (either 1 or 0) will be considered 1 only if all the neighbouring pixels of that pixel is 1, otherwise it is replaced with the value zero.This also eliminates any isolated small pixel groups present in the image. Working of erosion:

  1) A kernel(a matrix of odd size(3,5,7)) with specific values is convolved with our binarised image.
  2) The convolution will make sure that A pixel in the original image (either 1 or 0) will be considered 1 only if all the neighbouring pixels of that pixel is 1, otherwise it is replaced with the value zero.
  3) Thus causing the thickness or boldness of the foreground to reduce

- **closing filter :** Finally there is a algorithm in opencv that allows us to check for the total area covered by pixel groups in the background(1).If the area is lesser than a theshold that particular group or isolated pixel is removed from the image thus removing most of the noise in the image.

The preprocessing pipeline guarantees that the digitized inscriptions are optimized for subsequent individual popularity algorithms. This complete method not only addresses challenges associated with image high-quality but also sets the degree for accurate and efficient decoding of ancient inscriptions from temple partitions, thereby contributing notably to the renovation and understanding of cultural historical past.

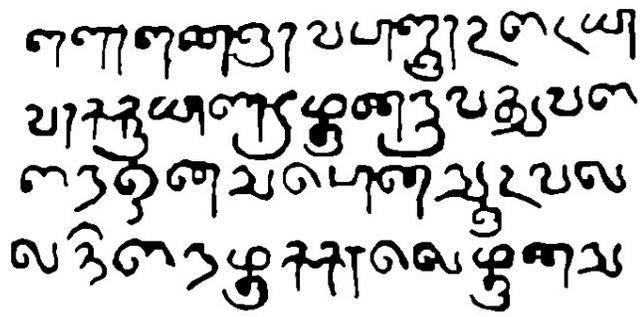

Fig. 7. Pre-processed image

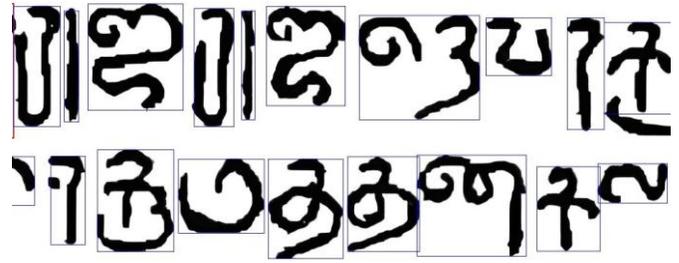

Fig. 8. Box rectification

## C. Data preparation

In image-based character recognition for temple inscriptions, the process of data preparation plays a pivotal role in refining the input data for subsequent training of Convolutional Neural Networks (CNNs). Following the preprocessing phase, the next vital step involves the predicting of bounding boxes using Tesseract using modern Tamil model and fine tuning the said model to be able to recognise ancient Tamil literature. The lack of groundtruth data lead us to create the ground truth by fine tuning modern Tamil data.The JTessBox Editor enables the creation and adjustment of bounding boxes around each distinct character inside the digitized inscriptions. This step is vital for appending the dataset with specific spatial records about the region and dimensions of every letter. Accurate bounding boxes serve as the inspiration for the CNN to research and understand patterns effectively, contributing to the overall efficiency of the laguage model. Manual intervention permits for the rectification of misalignment, ambiguous characters, or instances wherein automated methods may additionally fall quick because of the precise characteristics of temple inscriptions.

By meticulously addressing and rectifying prediction errors, the dataset will become a precious useful resource for training the CNN, fostering a greater efficient and precise character reputation gadget. This meticulous facts practise phase, combining automated bounding field generation with manual fine-tuning, establishes a stable basis for the following education of the CNN, in the end enhancing the device's functionality to decode and interpret temple inscriptions with a high degree of accuracy and cultural importance.

## D. Tesseract OCR

In the context of character recognition for the Tamil font in our task, Tesseract OCR plays a pivotal position in effectively interpreting digitized temple inscriptions. The method starts off with the extraction of Unicharset characters.

The unicharset extractor application is utilized to create a unicharset file, capturing the particular characters position/coordinates and identity completely. Following this,the font properties is defined.which is used by the unicharset extractor to create .tr files.

In the training section, the intermediary files are fed to 'mf-training' application and then finally a cnntraining application 'cntraining'. These commands utilize the generated character files files to create important components like 'inttemp', 'normproto', 'pffmtable', and 'shapetable'. These components are then combined into a unified 'atam.Traineddata' using the 'combine tessdata' command.

Finally, the atam.traineddata is copied to the appropriate source folder under tessdata, permitting Tesseract OCR to access the custom-trained version for Tamil font recognition. This guarantees that Tesseract is ready with the essential knowledge to appropriately decipher characters present within the temple inscriptions.

## E. Post processing

In the post-processing phase of our image-based character recognition system for temple inscriptions, a critical step involves refining the output generated by Tesseract OCR. The raw output lacks the use of spaces between words and use of punctuation , which necessitates the implementation of a word break segmentation algorithm. This algorithm is instrumental in intelligently including spaces and pauses between applicable words, enhancing the general clarity and coherence of the deciphered inscriptions.

The word break segmentation algorithm systematically analyzes the string of characters, using contextual information in form dictionary of words to pick out appropriate locations for word breaks. This intelligent segmentation uses a binary heap for better performance with low run time.

By introducing spaces and pauses strategically, the post-processing section transforms the output into a coherent illustration of the temple inscriptions. The implementation of such post-processing strategies is vital for bridging the space between uncooked OCR output and a sophisticated, human-readable representation, ultimately enriching the value of our photograph-based totally person recognition system for temple inscriptions.

## III. RESULTS AND DISCUSSION

This work aims on contributing to the development of a system to digitise ancient scripts accurately and efficiently. Furthermore document the knowledge on ancient scriptures we learn from Tamil scholars. Numerous scanned images from various temples were obtained and pre-processed as a test

```
#driver code
for i in str:
    if i == ' ':
        i='\0'

# creating object of WordBreak() class
ob = WordBreak()
# displaying results
ans = ob.wordBreak(str,["உடையார","கோயிலில","திருமகள", "பெருநி", "ஓகச்", "செல்வியுளக்கே", "உரிமை", "பூண்ட_மை"])
print(ans)
```

Fig. 9. Post-processing code

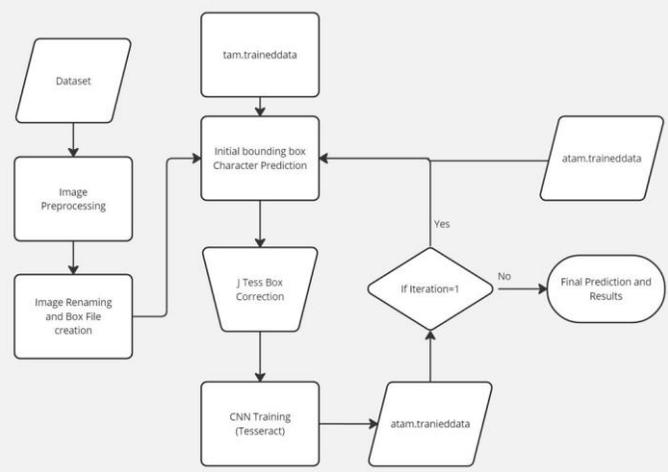

Fig. 10. Training Process

image for our model. wherein a large pool of images were obtained from Brihadeeshwarar temple. A llac kof dataset caused us to curate a moderately large dataset in 10th century Tamil from various temple inscriptions and handwritten text from our team and volunteers.we have used Jtess editor and Tesseract training tools to train our custom ancient Tamil scripture model using CNNs in Tesseract.further a web application was developed for the use of the technology using phone cameras or using pre-existing images.

### A. Custom model testing

The forementioned test images were pre-processsed appropriately and fed into our custom language model. Using training tools provided by tessearct OCR the required ground truth files were generated. After which they were trained in the Tesseract CNN environment using modern Tamil model as template to create our required model(traineddata file).

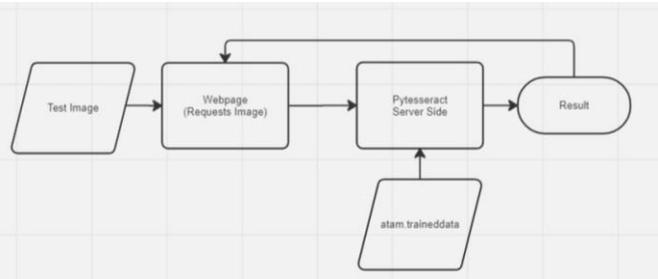

Fig. 11. Model Presentation

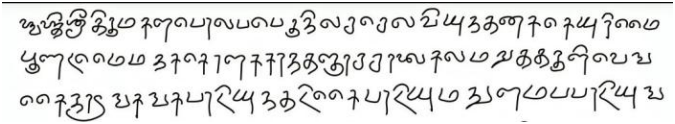

Fig. 12. Handwritten scripture written from Brihadeeswarar Temple

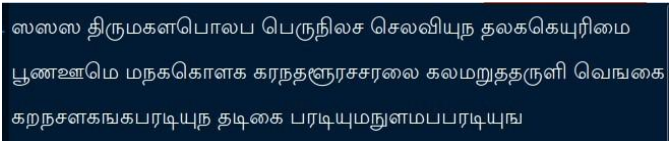

Fig. 13. Output obtained in the command line

After which these images were trained again but using the previously trained model as a template.This way our model would be more fine tuned towards predicting ancient Tamil texts. The accuracy was tested for 5 different images as test data and average accuracy of the model across them were taken. The finalised model took approximately 2 hrs to train its neural network.The finalised model was tested against 10 test images and in each of the images the average correct predictions rate was taken.A weighted average of each of these values gave us the performance of the model against the test images.Weighted average was taken since each of the test had non-uniform number of characters in them.Initially we observed very abysmal performance rates,which mostly was due to segmentation errors,To overcome this tesseract had options that could modify how the image is processed by tesseract to draw boxes,we changed this parameter to psm 6,which treats the miages as a pdf document with uniformly spaced characters in rows read from left to right. After which We observed a overall performance of 80.8% which is well within our expected range of error.In which we didnt take into account the prediction of punctuated characters,wherein any punctuated characters in the actual result required was rewritten as their base character.The reason for the decrease in performance mostly had to do with the morphological characters of Tamil characters that consist of multi-character compound single letters and also had to do with some characters that had their curves over the adjacent characters essentially creating a very unique character differing from the norm.

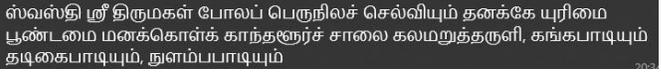

Fig. 14. The expected text for fig 11

### B. Web-application

Integration of our trained model with a dynamic web application was performed for public ease of use and presentation.The tools used for the creation of the website were flask api,pytesseract and our trained data.The server side processing of the image is done by first pre-processing the images into the format required by our model,and then the image is fed into our tesseract model which extracts the text from our image.Finally this extracted text is sent back to the website for display.

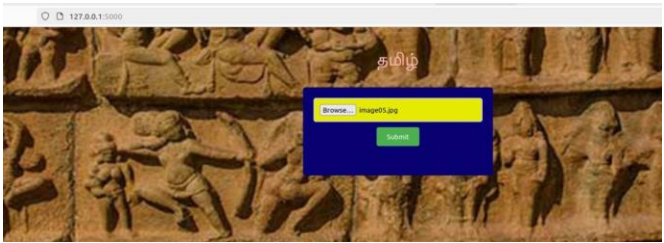

Fig. 15. submission page

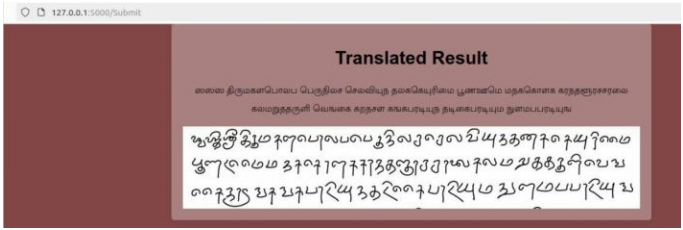

Fig. 16. translated output

The model was implemented on a flask web server using python.Python has an tesseract api that enables us to use our tesseract in very short lines of code without the use of os api.Below is the code snippet used to predict with our custom model which we named atam.

## IV. CHALLENGES FACED

1) The lack of spaces between words in the scriptures required us to create a separate algorithm to tackle this problem and required a database of the words used in the scripture.
2) Each word could mean a lot of things due to the absence of punctuation's and the difference between the auxiliary character in 'aah' sound and the 'ra' character when written in ancient tamil.
3) Each image taken was unique in its lighting angling due to how and where its written.Hence a universal pre-processing code applicable to all the images taken wasn't able to be made.some image underwent good processing in different values of window constant and threshold value.
4) Tesseract bounding box algorithm is based on latin scripts that lacks the use of compund characters unlike tamil thus the model draws a box for each single character in the compund letter.
5) Thus a language specific bounding box algorithm or a separate neural network was needed to be trained for drawing boxes properly.
6) Most characters had overlapping bounding boxes that disrupted the prediction process,This was solved by localising the boxes to a portion of the letter uniformly for all occurances of that letter

## V. CONCLUSION

In this work we assembled a dataset of ancient tamil charaters using various means and documented the same.this work also attemted to create a custom language model in tesseract ocr for ancient tamil scriptures,By using the inbuilt lstm network in tesseract and opencv to process the images, As discussed previous numerous difficulties in the way the ancient letters are written and the vast difference from modern tamil led to errors due to segmentation.As a future work a separate segmentation/bounding box generating AI model needs to be procured to further increase the accuracy. The final accuracy of our model was within the acceptable range of 80.8%.The efficiency can be further improved by expanding the dataset even more. There is also scope to to try and implement ICR technologies to obtain even more thoroughly and accurately pre-processed images for Tesseract to use.In terms of adding in punctuation into the image a need for extensive research and most likely a AI system that is trained on the occurrence of 'mei eluthukkal' in modern tamil words are required.